\documentclass[11pt,a4paper]{article}
\usepackage[hyperref]{naaclhlt2019}
\usepackage{times}
\usepackage{latexsym}
\usepackage{amsmath}
\usepackage{amssymb}
\usepackage{booktabs}
\usepackage{multirow}
\usepackage{url}
\usepackage{graphicx}

\aclfinalcopy 

\title{BERT has a Mouth, and It Must Speak: \\
       BERT as a Markov Random Field Language Model}

\author{Alex Wang \\
  New York University \\
  \texttt{alexwang@nyu.edu} \\\And
  Kyunghyun Cho \\
  New York University \\ 
  Facebook AI Research  \\
  CIFAR Azrieli Global Scholar \\
  \texttt{kyunghyun.cho@nyu.edu} \\}

\date{}

\begin{document}
\maketitle
\begin{abstract}
We show that BERT \citep{devlin2018bert} is a Markov random field language model. 
This formulation gives way to a natural procedure to sample sentences from BERT. 
We generate from BERT and find that it can produce high-quality, fluent generations. 
Compared to the generations of a traditional left-to-right language model, BERT generates sentences that are more diverse but of slightly worse quality.
\end{abstract}

\section{Introduction}

BERT~\citep{devlin2018bert} is a recently released sequence model used to achieve state-of-art results on a wide range of natural language understanding tasks, including constituency parsing \citep{kitaev2018multilingual} and machine translation \citep{lample2019cross}.
Early work probing BERT's linguistic capabilities has found it surprisingly robust \citep{goldberg2019assessing}.

BERT is trained on a \textit{masked language modeling} objective.
Unlike a traditional language modeling objective of predicting the next word in a sequence given the history, masked language modeling predicts a word given its left and right context.
Because the model expects context from both directions, it is not immediately obvious how BERT can be used as a traditional language model (i.e., to evaluate the probability of a text sequence) or how to sample from it.

We attempt to answer these questions by showing that BERT is a combination of a Markov random field language model~\citep[MRF-LM,][]{jernite2015fast,mikolov2013efficient} with pseudo log-likelihood~\citep{besag1977efficiency} training. 
This formulation automatically leads to a sampling procedure based on Gibbs sampling.

\section{BERT as a Markov Random Field}

Let $X=(x_1, \ldots, x_T)$ be a sequence of random variables $x_i$, each of which is categorical in that it can take one of $M$ items from a vocabulary $V=\left\{ v_1, \ldots, v_{M} \right\}$. These random variables form a fully-connected graph with undirected edges, indicating that each variable $x_i$ is dependent on all the other variables.

\paragraph{Joint Distribution}

To define a Markov random field (MRF), we start by defining a potential over cliques. Among all possible cliques, we only consider the clique corresponding to the full graph. All other cliques will be assigned a potential of $1$ (i.e. $\exp(0)$). The potential for this full-graph clique decomposes into 
a sum of $T$ $\log$-potential terms:
\begin{align*}
    \phi(X) = \prod_{t=1}^T \phi_t(X) = \exp\left( \sum_{t=1}^T \log \phi_t(X) \right),
\end{align*}
where we use $X$ to denote the fully-connected graph created from the original sequence. Each log-potential $\phi_t(X)$ is defined as
\begin{align}
\label{eq:log-potential}
    \log \phi_t(X) = 
    \begin{cases}
    \text{1h}(x_t)^\top f_{\theta}(X_{\backslash t}),&\text{if } \left[ \text{MASK} \right]\notin \\
    & 
    \quad X_{1:t-1} \cup X_{t+1:T}  \\
    0, & \text{otherwise},
    \end{cases}
\end{align}
where $f_{\theta}(X_{\backslash t}) \in \mathbb{R}^M$, 
$\text{1h}(x_t)$ is a one-hot vector with index $x_t$ set to 1,
and
\[
X_{\backslash t}=(x_1, \ldots, x_{t-1}, \left[ \text{MASK} \right], x_{t+1}, \ldots, x_T)
\]
From this log-potential, we can define a probability of a given sequence $X$ as
\begin{align}
\label{eq:p-unnorm}
    p_{\theta}(X) = \frac{1}{Z(\theta)} \prod_{t=1}^T \phi_t(X),
\end{align}
where 
\[
Z(\theta) = \sum_{X'} \prod_{t=1}^T \phi_t(X'),
\]
for all $X'$. This normalization constant is unfortunately impractical to compute exactly, rendering exact maximum log-likelihood intractable. 

\paragraph{Conditional Distribution}

Given a fixed $X_{\backslash t}$, the conditional probability of $x_t$ is derived to be
\begin{align}
\label{eq:p-cond}
    p(x_t | X_{\backslash t}) = \frac{1}{Z(X_{\backslash t})} \exp(\text{1h}(x_t)^\top f_{\theta}(X_{\backslash t})),
\end{align}
where
\[
Z(X_{\backslash t}) = \sum_{m=1}^M \exp(\text{1h}(m)^\top f_{\theta}(X_{\backslash t})).
\]
This derivation follows from the peculiar formulation of the log-potential in Eq.~\eqref{eq:log-potential}. It is relatively straightforward to compute, as it is simply softmax normalization over $M$ terms~\citep{bridle1990probabilistic}. 

\paragraph{(Stochastic) Pseudo Log-Likelihood Learning}

One way to avoid the issue of intractability in computing the normalization constant $Z(\theta)$ above\footnote{
    In BERT it is not intractable in the strictest sense, since the amount of computation is bounded (by $T=500$) each iteration. It however requires computation up to $\exp(500)$ which is in practice impossible to compute exactly. 
}
is to resort to an approximate learning strategy. BERT uses pseudo log-likelihood learning, where the pseudo log-likelihood is defined as:
\begin{align}
\label{eq:pll}
    \text{PLL}(\theta; D) = 
    \frac{1}{|D|} \sum_{X \in D}
    \sum_{t=1}^{|X|}
    \log p(x_t | X_{\backslash t}),.
\end{align}
where $D$ is a set of training examples. We maximize the predictability of each token in a sequence given all the other tokens, instead of the joint probability of the entire sequence. 

It is still expensive to compute the pseudo log-likelihood in Eq.~\eqref{eq:pll} for even one example, especially when $f_{\theta}$ is not linear. This is because we must compute $|X|$ forward passes of $f_{\theta}$ for each sequence, when $|X|$ can be long and $f_{\theta}$ be computationally heavy. Instead we could stochastically estimate it by
\begin{align*}
\frac{1}{|X|}
    \sum_{t=1}^{|X|}
    \log p& (x_t | X_{\backslash t})
    \\
    =&
    \mathbb{E}_{t \sim \mathcal{U}(\left\{ 1,\ldots, |X|\right\})}
    \left[
    \log p(x_t | X_{\backslash t})
    \right]
    \\
    \approx&
    \frac{1}{K} \sum_{k=1}^K
    \log p(x_{\tilde{t}_k} | X_{\backslash \tilde{t}_k}),
\end{align*}
where $\tilde{t}_k \sim \mathcal{U}(\left\{ 1,\ldots, |X|\right\}$. Let us refer to this as stochastic pseudo log-likelihood learning.

\paragraph{In Reality}

The stochastic pseudo log-likelihood learning above states that we ``mask out'' one token in a sequence at a time and let $f_{\theta}$ predict it based on all the other ``observed'' tokens in the sequence. \citet{devlin2018bert} however proposed to ``mask out'' multiple tokens at a time and predict all of them given both all ``observed'' and ``masked out'' tokens in the sequence. This brings the original BERT closer to a denoising autoencoder~\citep{vincent2010stacked}, which could still be considered as training a Markov random field with (approximate) score matching~\citep{vincent2011connection}.

\begin{table*}[h]
    \centering
    \small
    \begin{tabular}{p{.48\textwidth}p{.48\textwidth}}
         \toprule
         the nearest regional centre is alemanno , with another connection to potenza and maradona , and the nearest railway station is in bergamo , where the line terminates on its northern end & for all of thirty seconds , she was n't going to speak . maybe this time , she 'd actually agree to go . thirty seconds later , she 'd been speaking to him in her head every \\
         \midrule
         ' let him get away , mrs . nightingale . you could do it again . ' ' he - ' ' no , please . i have to touch him . and when you do , you run . & `` oh , i 'm sure they would be of a good service , '' she assured me . `` how are things going in the morning ? is your husband well ? '' `` yes , very well \\
         \midrule
         he also `` turned the tale [ of ] the marriage into a book '' as he wanted it to `` be elegiac '' . both sagas contain stories of both couple and their wedding night ; & `` i know . '' she paused .`` did he touch you ? '' `` no . '' `` ah . '' `` oh , no , '' i said , confused , not sure why \\
         \midrule
         `` i had a bad dream . '' `` about an alien ship ? who was it ? '' i check the text message that 's been only partially restored yet, the one that says love .  & i watched him through the glass , wondering if he was going to attempt to break in on our meeting . but he did n't seem to even bother to knock when he entered the room . i was n't \\
         \midrule
         replaced chris hall ( st . louis area manager ) . june 9 : mike howard ( syndicated `` good morning '' , replaced steve koval , replaced dan nickolas , and replaced phil smith ) ; &  `` how long has it been since you have made yourself an offer like that ? '' asked planner . `` oh '' was the reply . planner had heard of some of his other business associates who had \\
         \bottomrule
    \end{tabular}
    \caption{Random sample generations from BERT base (left) and GPT (right).}
    \label{tab:generations}
\end{table*}

\section{Using BERT as an MRF-LM}\label{sec:sampling}

The discussion so far implies that BERT is a Markov random field language model (MRF-LM) and that it learns a distribution over sentences (of some given length). This framing suggests that we can use BERT not only as parameter initialization for finetuning but as a generative model of sentences to either score a sentence or sample a sentence. 

\paragraph{Ranking}

Let us fix the length $T$. Then, we can use BERT to rank a set of sentences. We cannot compute the exact probabilities of these sentences, but we can compute their unnormalized log-probabilities according to Eq.~\eqref{eq:p-unnorm}:
\[
\sum_{t=1}^T \log \phi_t(X).
\]
These unnormalized probabilities can be used to find the most likely sentence within the set or to sort the sentences according to their probabilities.

\paragraph{Sampling}

Sampling from a Markov random field is less trivial than is from a directed graphical model which naturally admits ancestral sampling. One of the most widely used approaches is Markov-chain Monte-Carlo (MCMC) sampling~\citep{neal1993probabilistic,swendsen1986replica,salakhutdinov2009learning,desjardins2010tempered,cho2010parallel}. In this report, we only consider Gibbs sampling which fits naturally with (stochastic) pseudo log-likelihood learning. 

In Gibbs sampling, we start with a random initial state $X^0$, which we initialize to be an all-mask sequence, i.e., $(\left[ \text{MASK} \right], \ldots, \left[ \text{MASK} \right])$, though we could with a sentence consisting of randomly sampled words or by retrieving a sentence from data.
At each iteration $i$, 
we sample the position $t^i$ uniformly at random from $\left\{1, \ldots, T\right\}$ and mask out the selected location, i.e., $x_{t^i}^i=\left[ \text{MASK} \right]$, resulting in $X^i_{\backslash t^i}$. We now compute $p(x_{t^i} | X^i_{\backslash t^i})$ according to Eq.~\eqref{eq:p-cond}, 
sample $\tilde{x}_{t^i}$ from it\footnote{
In practice, one can imagine sampling from the $k$-most probable words~\citep{fan2018hierarchical}.
We find $k=100$ to be effective in early experiments.}, and construct the next sequence by
\begin{align*}
    X^{i+1} = (x^i_1, \ldots, x^i_{t^i-1}, \tilde{x}_{t^i}, 
    x^i_{t^i+1}, \ldots, x^i_T).
\end{align*}
We repeat this procedure many times, preferably with thinning.\footnote{
Thinning refers to the procedure of selecting a sample only once a while during MCMC sampling.
}
Because Gibbs sampling, as well as any MCMC sampler with a local proposal distribution, tends to get stuck in a mode of the distribution, we advise running multiple chains of Gibbs sampling or using different sentence initializations.

\paragraph{Sequential Sampling}

The undirectedness of the MRF-LM and the bidirectional nature of BERT do not naturally admit sequential sampling, but given that the dominant approach to text generation is left-to-right, we experiment with generating from BERT in such a manner.

As with our non-sequential sampling scheme, we can begin with a seed sentence of either all masks or a random sentence.
Whereas previously we sampled a position $t \in \{1, \dots, T \}$ to mask out and generate for at each time step, in the sequential setting, at each time step $t$, we mask out $x^t_t$, generate a word for that position, and substitute it into the sequence. After $T$ timesteps, we have a sampled a token at each position, at which we point we can terminate or repeat the process from the current sentence. 



\begin{table*}[t]
    \footnotesize
    \centering
    \begin{tabular}{ccrrrrrrrrr}
         \toprule
         \multirow{3}{*}{Model} & \multirow{3}{*}{Self-BLEU ($\downarrow$)} & \multicolumn{9}{c}{\% Unique $n$-grams ($\uparrow$)} \\ \cmidrule{3-11}
         & & \multicolumn{3}{c}{Self} & \multicolumn{3}{c}{WT103} & \multicolumn{3}{c}{TBC} \\ \cmidrule{3-5} \cmidrule{6-8} \cmidrule{9-11}
         & & n=2 & n=3 & n=4 & n=2 & n=3 & n=4 & n=2 & n=3 & n=4 \\
         \midrule
         BERT (large) & 9.43 & 63.15 & 92.38 & 98.01 & 59.91 & 91.86 & 98.43 & 64.59 & 93.27 & 98.59 \\
         BERT (base) & 10.06 & 60.76 & 91.76 & 98.14 & 57.90 & 91.72 & 98.55 & 60.94 & 92.04 & 98.56 \\
         GPT & 40.02 & 31.13 & 67.01 & 87.28 & 33.71 & 72.86 & 91.12 & 25.74 & 65.04 & 88.42 \\
         WT103 & 9.80 & 70.29 & 94.36 & 99.05 & 56.19 & 88.05 & 97.44 & 68.35 & 94.20 & 99.23 \\
         TBC & 12.51 & 62.19 & 92.70 & 98.73 & 55.30 & 91.08 & 98.81 & 44.75 & 82.06 & 96.31 \\
         \bottomrule
    \end{tabular}
    \caption{Self-BLEU and percent of generated $n$-grams that are unique relative to own generations (left) WikiText-103 test set (middle) a sample of 5000 sentences from Toronto Book Corpus (right). For the WT103 and TBC rows, we sample 1000 sentences from the respective datasets. }
    \label{tab:ngrams}
\end{table*}

\section{Experiments}

Our experiments demonstrate the potential of using BERT as a {\it standalone} language model rather than as a parameter initializer for transfer learning~\citep{devlin2018bert,lample2019cross,nogueira2019passage}. 
We show that sentences sampled from BERT are well-formed and are assigned high probabilities by an off-the-shelf language model.
We take pretrained BERT models trained on a mix of Toronto Book Corpus \citep[TBC, ][]{moviebook} and Wikipedia provided by \citet{devlin2018bert} and its PyTorch implementation\footnote{
\url{https://github.com/huggingface/pytorch-pretrained-BERT}
}  provided by HuggingFace. We experiment with both the base and large BERT configuations.

\subsection{Evaluation}

We consider several evaluation metrics to estimate the quality and diversity of the generations.

\paragraph{Quality} To automatically measure the quality of the generations, we follow \citet{yu2017seqgan} by computing BLEU \citep{papineni2002bleu} between the generations and the original data distributions to measure how similar the generations are.
We use a random sample of 5000 sentences from the test set of WikiText-103 \citep[WT103,  ][]{merity2016pointer} and a random sample of 5000 sentences from TBC as references.

We also use the perplexity of a trained language model evaluated on the generations as a rough proxy for fluency. Specifically, we use the Gated Convolutional Language Model \citep{dauphin2016language} pretrained on WikiText-103\footnote{\url{https://github.com/pytorch/fairseq/tree/master/examples/conv_lm}}.

\paragraph{Diversity} To measure the diversity of each model's generations, we compute self-BLEU \citep{zhu2018texygen}: for each generated sentence, we compute BLEU treating the rest of the sentences as references, and average across sentences. Self-BLEU measures how similar each generated sentence is to the other generations; high self-BLEU indicates that the model has low sample diversity.

We also evaluate the percentage of $n$-grams that are unique, when compared to the original data distribution and within the corpus of generations. We note that this metric is somewhat in opposition to BLEU between generations and data, as fewer unique $n$-grams implies higher BLEU.

\paragraph{Methodology} We use the non-sequential sampling scheme with sampling from the top $k=100$ most frequent words at each time step, as empirically this led to the most coherent generations. We show generations from the sequential sampler in Table \ref{tab:more_generations} in the appendix.
We compare against generations from a high-quality neural language model, the OpenAI Generative Pre-Training Transformer \citep[GPT]{radford2018improving}, which was trained on TBC and has approximately the same number of parameters as the base configuration of BERT. 
For BERT, we pad each input with special symbols $\left[ \text{CLS} \right]$ and $\left[ \text{SEP} \right]$. 
For GPT, we start with a start of sentence token and generate left to right.
For all models, we generate 1000 uncased sequences of length 40.
Finally, as a trivial baseline, we sample 1000 sentences from TBC and the training split of WT103 and compute all automatic metrics against these samples.

\begin{table}[t]
    \footnotesize
    \centering
    \begin{tabular}{cccc}
         \toprule
         \multirow{2}{*}{Model} & \multicolumn{2}{c}{Corpus-BLEU ($\uparrow$)} & \multirow{2}{*}{PPL ($\downarrow$)} \\ \cmidrule{2-3}
         & WT103 & TBC & \\
         \midrule
         BERT (large) & 5.05 & 7.60 & 331.47 \\
         BERT (base) & 7.80 & 7.06 & 279.10 \\
         GPT & 10.81 & 30.75 & 154.29 \\
         WT103  & 17.48 & 6.57 & 54.00 \\
         TBC & 10.05 & 23.05 & 314.28 \\
         \bottomrule
    \end{tabular}
    \caption{Quality metrics of model generations. Perplexity (PPL) is measured using an additional language model \citep{dauphin2016language}. For the WT103 and TBC rows, we sample 1000 sentences from the respective datasets.}
    \label{tab:results}
\end{table}

\begin{figure*}
    \centering
    \includegraphics[width=.90\linewidth]{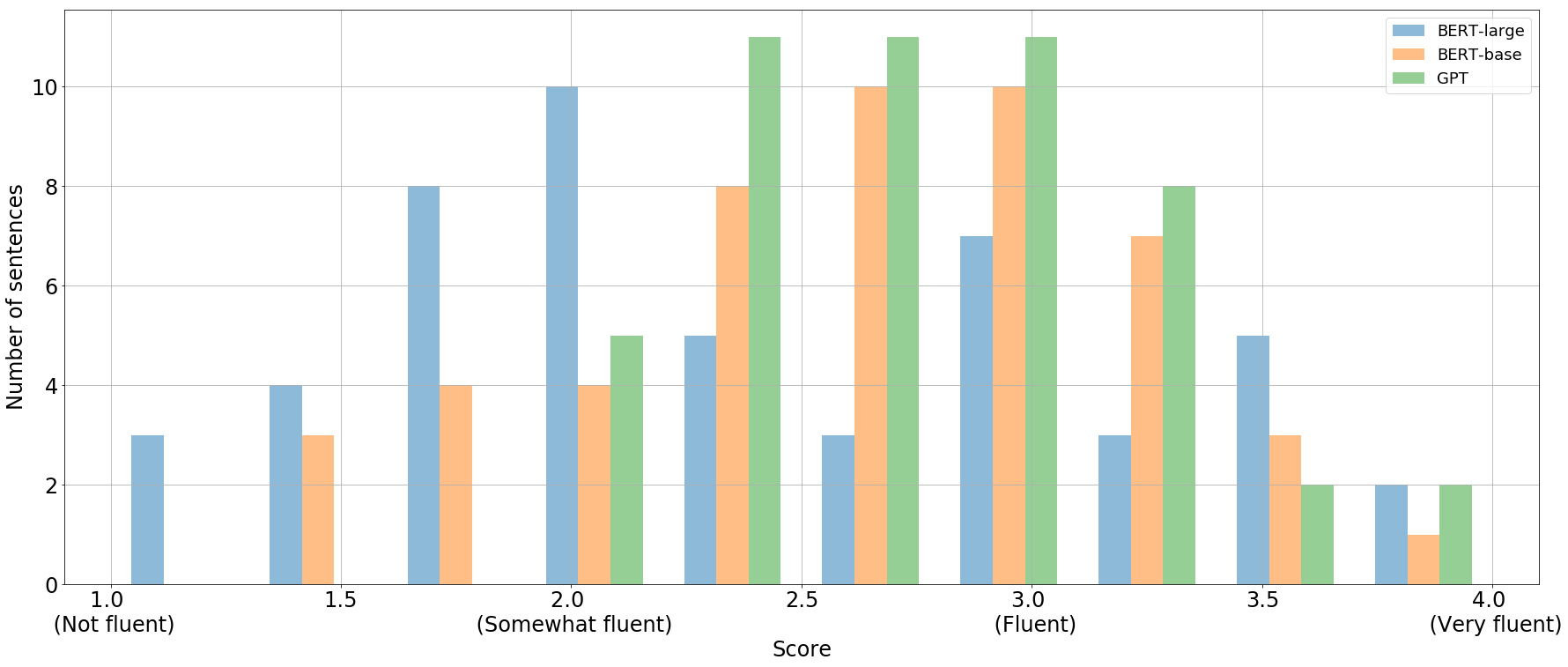}
    \caption{Fluency scores for 100 sentences samples from each of BERT large, BERT base, and GPT, as judged by human annotators according to a four-point Likert scale.}
    \label{fig:human}
\end{figure*}

\section{Results}

We present sample generations, quality results, and diversity results respectively in Tables \ref{tab:generations}, \ref{tab:ngrams}, \ref{tab:results}.

We find that, compared to GPT, the BERT generations are of worse quality, but are more diverse. 
Surprisingly, the outside language model, which was trained on Wikipedia, is less perplexed by the GPT generations than the BERT generations, even though GPT was only trained on romance novels and BERT was trained on romance novels and Wikipedia. 
On actual data from TBC, the outside language model is about as perplexed as on the BERT generations, which suggests that domain shift is an issue in using a trained language model for evaluating generations and that the GPT generations might have collapsed to fairly generic and simple sentences.
This observation is further bolstered by the fact that the GPT generations have a higher corpus-BLEU with TBC than TBC has with itself.
The perplexity on BERT samples is not absurdly high, and in reading the samples, we find that many are fairly coherent.
The corpus-BLEU between BERT models and the datasets is low, particularly with WT103.

We find that BERT generations are more diverse than GPT generations. GPT has high $n$-gram overlap (smaller percent of unique $n$-grams) with TBC, but surprisingly also with WikiText-103, despite being trained on different data. Furthermore, GPT generations have greater $n$-gram overlap with these datasets than these datasets have with themselves, further suggesting that GPT is relying significantly on generic sentences. BERT has lower $n$-gram overlap with both corpora, with similar degrees of $n$-gram overlap as the samples of the data.

For a more rigorous evaluation of generation quality, we collect human judgments on sentence fluency for 100 samples from BERT large, BERT base, and GPT using a four point Likert scale. For each sample we ask three annotators to rate the sentence on its fluency and take the average of the three judgments as the sentence's fluency score. 
We present a histogram of the results in Figure~\ref{fig:human}.
For BERT large, BERT base, and GPT we respectively get mean scores over the samples of 2.37 ($\sigma = 0.83$), 2.65 $(\sigma = 0.65)$, and 2.80 ($\sigma = 0.51$). All means are within a standard deviation of each other. BERT base and GPT have similar unimodal distributions with BERT base having a slightly more non-fluent samples. BERT large has a bimodal distribution. 

\section{Conclusion}

We show that BERT is a Markov random field language model. Formulating BERT in this way gives rise to a practical algorithm for generating from BERT based on Gibbs sampling that does not require any additional parameters or training. We verify in experiments that the algorithm produces diverse and fairly fluent generations. 
The power of this framework is in allowing the principled application of Gibbs sampling, and potentially other MCMC algorithms, for generating from BERT.

Future work might explore these improved sampling methods, especially those that do not need to run the model over the entire sequence each iteration and that more robustly handle variable-length sequences.
To facilitate such investigation, we release our code on GitHub at \url{https://github.com/nyu-dl/bert-gen} and a demo as a Colab notebook at \url{https://colab.research.google.com/drive/1MxKZGtQ9SSBjTK5ArsZ5LKhkztzg52RV}.

\section*{Acknowledgements}
We thank Ilya Kulikov and Nikita Nangia for their help, as well as reviewers for insightful comments.
AW is supported by an NSF Fellowship. KC is partly supported by Samsung Advanced Institute of Technology (Next Generation Deep Learning: from Pattern Recognition to AI) and Samsung Electronics (Improving Deep Learning using Latent Structure).

\bibliography{naaclhlt2019}

\begin{thebibliography}{24}
\expandafter\ifx\csname natexlab\endcsname\relax\def\natexlab#1{#1}\fi

\bibitem[{Besag(1977)}]{besag1977efficiency}
Julian Besag. 1977.
\newblock Efficiency of pseudolikelihood estimation for simple gaussian fields.
\newblock \emph{Biometrika}, pages 616--618.

\bibitem[{Bridle(1990)}]{bridle1990probabilistic}
John~S Bridle. 1990.
\newblock Probabilistic interpretation of feedforward classification network
  outputs, with relationships to statistical pattern recognition.
\newblock In \emph{Neurocomputing}, pages 227--236. Springer.

\bibitem[{Cho et~al.(2010)Cho, Raiko, and Ilin}]{cho2010parallel}
KyungHyun Cho, Tapani Raiko, and Alexander Ilin. 2010.
\newblock Parallel tempering is efficient for learning restricted boltzmann
  machines.
\newblock In \emph{Neural Networks (IJCNN), The 2010 International Joint
  Conference on}, pages 1--8. IEEE.

\bibitem[{Dauphin et~al.(2016)Dauphin, Fan, Auli, and
  Grangier}]{dauphin2016language}
Yann~N Dauphin, Angela Fan, Michael Auli, and David Grangier. 2016.
\newblock Language modeling with gated convolutional networks.
\newblock \emph{arXiv preprint:1612.08083}.

\bibitem[{Desjardins et~al.(2010)Desjardins, Courville, Bengio, Vincent, and
  Delalleau}]{desjardins2010tempered}
Guillaume Desjardins, Aaron Courville, Yoshua Bengio, Pascal Vincent, and
  Olivier Delalleau. 2010.
\newblock Tempered markov chain monte carlo for training of restricted
  boltzmann machines.
\newblock In \emph{Proceedings of the thirteenth international conference on
  artificial intelligence and statistics}, pages 145--152.

\bibitem[{Devlin et~al.(2018)Devlin, Chang, Lee, and
  Toutanova}]{devlin2018bert}
Jacob Devlin, Ming-Wei Chang, Kenton Lee, and Kristina Toutanova. 2018.
\newblock Bert: Pre-training of deep bidirectional transformers for language
  understanding.
\newblock \emph{arXiv preprint:1810.04805}.

\bibitem[{Fan et~al.(2018)Fan, Lewis, and Dauphin}]{fan2018hierarchical}
Angela Fan, Mike Lewis, and Yann Dauphin. 2018.
\newblock Hierarchical neural story generation.
\newblock \emph{arXiv preprint:1805.04833}.

\bibitem[{Goldberg(2019)}]{goldberg2019assessing}
Yoav Goldberg. 2019.
\newblock \href {http://arxiv.org/abs/1901.05287} {Assessing bert's syntactic
  abilities}.

\bibitem[{Jernite et~al.(2015)Jernite, Rush, and Sontag}]{jernite2015fast}
Yacine Jernite, Alexander Rush, and David Sontag. 2015.
\newblock A fast variational approach for learning markov random field language
  models.
\newblock In \emph{International Conference on Machine Learning}, pages
  2209--2217.

\bibitem[{Kitaev and Klein(2018)}]{kitaev2018multilingual}
Nikita Kitaev and Dan Klein. 2018.
\newblock \href {http://arxiv.org/abs/1812.11760} {Multilingual constituency
  parsing with self-attention and pre-training}.

\bibitem[{{Lample} and {Conneau}(2019)}]{lample2019cross}
Guillaume {Lample} and Alexis {Conneau}. 2019.
\newblock {Cross-lingual Language Model Pretraining}.
\newblock \emph{arXiv e-prints}, page arXiv:1901.07291.

\bibitem[{Merity et~al.(2016)Merity, Xiong, Bradbury, and
  Socher}]{merity2016pointer}
Stephen Merity, Caiming Xiong, James Bradbury, and Richard Socher. 2016.
\newblock \href {http://arxiv.org/abs/1609.07843} {Pointer sentinel mixture
  models}.

\bibitem[{Mikolov et~al.(2013)Mikolov, Chen, Corrado, and
  Dean}]{mikolov2013efficient}
Tomas Mikolov, Kai Chen, Greg Corrado, and Jeffrey Dean. 2013.
\newblock Efficient estimation of word representations in vector space.
\newblock \emph{arXiv preprint:1301.3781}.

\bibitem[{Neal(1993)}]{neal1993probabilistic}
Radford~M Neal. 1993.
\newblock Probabilistic inference using markov chain monte carlo methods.

\bibitem[{Nogueira and Cho(2019)}]{nogueira2019passage}
Rodrigo Nogueira and Kyunghyun Cho. 2019.
\newblock Passage re-ranking with bert.
\newblock \emph{arXiv preprint:1901.04085}.

\bibitem[{Papineni et~al.(2002)Papineni, Roukos, Ward, and
  Zhu}]{papineni2002bleu}
Kishore Papineni, Salim Roukos, Todd Ward, and Wei-Jing Zhu. 2002.
\newblock Bleu: a method for automatic evaluation of machine translation.
\newblock In \emph{Proceedings of the 40th annual meeting on association for
  computational linguistics}, pages 311--318. Association for Computational
  Linguistics.

\bibitem[{Radford et~al.(2018)Radford, Narasimhan, Salimans, and
  Sutskever}]{radford2018improving}
Alec Radford, Karthik Narasimhan, Tim Salimans, and Ilya Sutskever. 2018.
\newblock Improving language understanding by generative pre-training.
\newblock \emph{URL https://s3-us-west-2. amazonaws.
  com/openai-assets/research-covers/languageunsupervised/language understanding
  paper. pdf}.

\bibitem[{Salakhutdinov(2009)}]{salakhutdinov2009learning}
Ruslan~R Salakhutdinov. 2009.
\newblock Learning in markov random fields using tempered transitions.
\newblock In \emph{Advances in neural information processing systems}, pages
  1598--1606.

\bibitem[{Swendsen and Wang(1986)}]{swendsen1986replica}
Robert~H Swendsen and Jian-Sheng Wang. 1986.
\newblock Replica monte carlo simulation of spin-glasses.
\newblock \emph{Physical review letters}, 57(21):2607.

\bibitem[{Vincent(2011)}]{vincent2011connection}
Pascal Vincent. 2011.
\newblock A connection between score matching and denoising autoencoders.
\newblock \emph{Neural computation}, 23(7):1661--1674.

\bibitem[{Vincent et~al.(2010)Vincent, Larochelle, Lajoie, Bengio, and
  Manzagol}]{vincent2010stacked}
Pascal Vincent, Hugo Larochelle, Isabelle Lajoie, Yoshua Bengio, and
  Pierre-Antoine Manzagol. 2010.
\newblock Stacked denoising autoencoders: Learning useful representations in a
  deep network with a local denoising criterion.
\newblock \emph{Journal of machine learning research}, 11(Dec):3371--3408.

\bibitem[{Yu et~al.(2017)Yu, Zhang, Wang, and Yu}]{yu2017seqgan}
Lantao Yu, Weinan Zhang, Jun Wang, and Yong Yu. 2017.
\newblock Seqgan: Sequence generative adversarial nets with policy gradient.
\newblock In \emph{AAAI}, pages 2852--2858.

\bibitem[{Zhu et~al.(2018)Zhu, Lu, Zheng, Guo, Zhang, Wang, and
  Yu}]{zhu2018texygen}
Yaoming Zhu, Sidi Lu, Lei Zheng, Jiaxian Guo, Weinan Zhang, Jun Wang, and Yong
  Yu. 2018.
\newblock Texygen: A benchmarking platform for text generation models.
\newblock \emph{arXiv preprint:1802.01886}.

\bibitem[{Zhu et~al.(2015)Zhu, Kiros, Zemel, Salakhutdinov, Urtasun, Torralba,
  and Fidler}]{moviebook}
Yukun Zhu, Ryan Kiros, Richard Zemel, Ruslan Salakhutdinov, Raquel Urtasun,
  Antonio Torralba, and Sanja Fidler. 2015.
\newblock Aligning books and movies: Towards story-like visual explanations by
  watching movies and reading books.
\newblock In \emph{arXiv preprint:1506.06724}.

\end{thebibliography}
\bibliographystyle{acl_natbib}

\appendix

\section{Other Sampling Strategies}

We explored two other sampling strategies: left-to-right and generating for all positions at each time step. See Section \ref{sec:sampling} for an explanation of the former. For the latter, we start with an initial sequence of all masks, and at each time step, we would not mask any positions but would generate for all positions. This strategy is designed to save on computation. However, we found that this tended to get stuck in non-fluent sentences that could not be recovered from. We present sample generations for the left-to-right strategy in Table \ref{tab:more_generations}. 

\begin{table*}[b]
    \centering
    \small
    \begin{tabular}{p{.95\textwidth}}
         \toprule
        all the good people , no more , no less . no more . for ... the kind of better people ... for ... for ... for ... for ... for ... for ... as they must become again . \\
         \midrule
sometimes in these rooms , here , back in the castle . but then : and then , again , as if they were turning , and then slowly , and and then and then , and then suddenly . \\
         \midrule
other available songs for example are the second and final two complete music albums among the highest played artists , including : the one the greatest ... and the last recorded album , " this sad heart " respectively . \\
         \midrule
6 that is i ? ? and the house is not of the lord . i am well ... the lord is ... ? , which perhaps i should be addressing : ya is then , of ye ? ? \\
         \midrule
four - cornered rap . big screen with huge screen two of his friend of old age . from happy , happy , happy . left ? left ? left ? right . left ? right . right ? ? \\
         \bottomrule
    \end{tabular}
    \caption{Random sample generations from BERT base using a sequential, left-to-right sampling strategy.}
    \label{tab:more_generations}
\end{table*}

\end{document}